\documentclass[preprint,sigconf]{acmart}
\renewcommand\footnotetextcopyrightpermission[1]{} % removes footnote with conference information in first column
\AtBeginDocument{%
  \providecommand\BibTeX{{%
    \normalfont B\kern-0.5em{\scshape i\kern-0.25em b}\kern-0.8em\TeX}}}

\setcopyright{none}
\makeatletter
\def\@copyrightspace{\relax}
\makeatother
\begin{document}

\title{Towards United Reasoning for Automatic Induction in Isabelle/HOL}

\author{Yutaka Nagashima}
%\authornote{
%This work was supported by the European Regional Development Fund under the project 
%AI \& Reasoning (reg. no.CZ.02.1.01/0.0/0.0/15\_003/0000466).}
\email{yutaka.nagashima@cvut.cz}
\orcid{0000-0001-6693-5325}
\affiliation{%
  \institution{CIIRC, Czech Technical University in Prague}
%  \city{Prague}
%  \country{Czech Republic}
}
\affiliation{%
  \institution{University of Innsbruck}
%  \city{Innsbruck}
%  \country{Austria}
}

\renewcommand{\shortauthors}{Yutaka Nagashima}

\begin{abstract}
  Inductive theorem proving is an important 
  long-standing challenge in computer science.
  In this extended abstract,
  we first summarize the recent developments of 
  proof by induction for Isabelle/HOL.
  Then, we propose \textit{united reasoning},
  a novel approach to further automating
  inductive theorem proving.
  Upon success,
  united reasoning 
  takes
  the best of three schools of reasoning:
  deductive reasoning, inductive reasoning,
  and inductive reasoning,
  to prove difficult inductive problems automatically.
\end{abstract}


\maketitle

\section{Induction in Isabelle/HOL}

Despite the importance of inductive theorem proving,
its automation remains as a long-standing challenge in computer science.
To facilitate proof by induction, 
Isabelle/HOL \cite{Isabelle} offers tools to apply induction,
such as the \verb|induct| method and 
the \verb|induction| method.

However, these tools were built for human-interaction 
rather than for automation.
For example,
when using the \verb|induct| method
proof authors are supposed to specify 
\begin{itemize}
    \item on which terms they apply induction,
    \item which variables they generalize, or
    \item which induction rules they use for recursion induction
\end{itemize}
\noindent
by passing arguments to the \verb|induct| method.

\section{PSL: Proof Strategy Language}
As our first step towards automatic induction,
we developed PSL \cite{psl}, 
a programmable, meta-tool framework for Isabelle/HOL.
Using PSL, one can specify the following 
example proof strategy:
\begin{verbatim}
strategy DInd = Thens [Dynamic(Induct), Auto, IsSolved]
\end{verbatim}
\noindent
When one applies this strategy to a given proof goal, 
PSL's runtime creates various \verb|induct| methods with arguments 
using the information within the proof goal
and its background proof context.
Then, it combines those \verb|induct| methods
with \verb|auto| and \verb|is_solved|, where 
\verb|auto| is the general purpose proof method,
and \verb|is_solved| checks if there is a remaining sub-goal or not.
Based on such combinations of methods,
PSL's runtime system executes an iterative deepening depth-first search,
trying to identify a combination of induction arguments, 
with which Isabelle can discharge the proof goal.

Note that
despite the confusing name
proof by induction is
an example of \textit{deductive reasoning},
as well as many other proof methods of Isabelle/HOL:
we can deduce the underlying induction principle
from the axioms of Isabelle/HOL.

\section{PGT: Goal-Oriented Conjecturing}

For some inductive problems
it is not enough for proof authors 
to apply the \verb|induct| method with arguments,
but they have to come up with useful auxiliary lemmas.
These auxiliary lemmas have to be strong enough
to derive the original goals,
but they should be provable at the same time.

We proposed our Proof Goal Transformer (PGT) \cite{pgt}
as an extension to PSL to facilitate such \textit{abductive reasoning}.

The advantage of this goal-oriented approach %to conjecturing 
is that
we can identify valuable conjectures 
out of many conjectures
through proof search
by combining the conjecturing mechanism with 
Isabelle's proof automation and counter-example finder.

For example, if PGT produces a conjecture, \verb|conjecture|,
while trying to prove an original goal, \verb|goal|,
PGT inserts \verb|conjecture| as an assumption of \verb|goal|,
transforming the original goal into the following two sub-goals:
\begin{itemize}
    \item \texttt{conjecture $\longrightarrow$ goal}
    \item \texttt{conjecture}
\end{itemize}
\noindent
Then, Isabelle attempts to discharge the first sub-goal
with a standard proof automation tool, \verb|fastforce|,
and attempts to refute the second sub-goal with
a counter-example finder, such as \verb|quickcheck|.
PGT discards this conjecture, \verb|conjecture|,
if \verb|fastforce| cannot discharge the first sub-goal
(\verb|conjecture| is not strong enough)
or
\texttt{quickcheck} finds a counter-example
(\verb|conjecture| is not provable).
This way, 
%even though most conjectures produced by PGT are not useful,
when combined with other sub-tools, %for Isabelle
PGT can focus on valuable conjectures
to avoid the explosion of the search space.

\section{MeLoId: Machine Learning Induction}

It is sometimes necessary to apply
nested induction 
with multiple conjecturing steps.
Unlike the aforementioned goal-oriented conjecturing approach,
there is no known way to narrow the search space
of the \verb|induct| method using counter-example finders:
the \verb|induct| method often transforms
a given proof goal into a base case and step cases
while preserving the provability
even when inappropriate arguments are given to the \verb|induct| method.

The sub-goals produced by the \verb|induct| method with inappropriate arguments 
are often still provable if the original goal is provable,
but are only harder to discharge 
because the inappropriate application of the \verb|induct| method transforms the original goals
to forms unsuitable for Isabelle's proof automation.

This blows up the search space of PSL
when attacking hard inductive problems
that require multiple applications of induction:
%This leads to the exponential blowup of search space
%when attacking hard inductive problems 
%that require multiple applications of induction with PSL:
each invocation of \texttt{Dynamic (Induct)} produces many
variants of \verb|induct| methods,
each of which produces inappropriate sub-goals, and
the nested applications of \texttt{Dynamic (Induct)}
produces again many inappropriate sub-goals for each
sub-goal produced by the previous application of the \verb|induct| method.

To keep the size of search space tractable,
we are currently developing MeLoId \cite{meloid}.
MeLoId is a supervised learning framework to suggest
promising arguments to the \verb|induct| method 
without completing a proof.

The overall architecture of MeLoId is similar to 
that of PaMpeR \cite{pamper},
which suggests promising proof methods for a given proof goal 
based on supervised learning on human-written proof corpora.
Upon success, 
MeLoId converts each invocation of 
the \verb|induct| method in proof corpora
and the corresponding proof goal into 
a list of boolean values.
Then, it applies a machine learning algorithms
to this simplified data and learns 
how human proof authors use the \verb|induct| method.

In 2018 we conducted our first small scale preliminary experiment
using around 40 feature extractors written in Isabelle/ML.
The result was not convincing:
even though the feature extractors did manage to distill the essence of some undesirable combinations of arguments for the \verb|induct| method,
they turned out to be unable to extract the essence of promising combinations of arguments for most cases.

The problem was that
for a feature extractor to distill induction heuristics,
such extractor has to be able to conduct a complex abstract reasoning
on different inductive problems across various problem domains.
Therefore,
it is harder to develop useful extractors for MeLoId directly in Isabelle/ML
than to develop the feature extractor used in PaMpeR.

For example,
a good feature extractor for MeLoId would reason
the syntactic structures of both the proof goals and 
the definitions of constants appearing in the goals
with respect to the induction terms passed as arguments to 
the \verb|induct| method,
whereas many feature extractors of PaMpeR
simply check the existence of atomic terms of certain names
within proof goals.

Due to this technical challenge,
we decided that
it is infeasible to 
develop useful feature extractors
for MeLoId in Isabelle/ML.
%To develop such complex abstract feature extractor for MeLoId,
And we developed a domain-specific language, LiFtEr \cite{lifter}, designed
to write feature extractors for MeLoId.
%so that we can develop feature extractors that can conduct aforementioned abstract complex reasoning.
LiFtEr allows experienced Isabelle users to encode their induction heuristics as assertions
in a style independent of any problem domain.

We plan to write many feature extractors in LiFtEr, 
extract a database from the Archive of Formal Proofs (AFPs) \cite{AFP} using those extractors,
and learn how Isabelle experts choose arguments for the \verb|induct| method,
so that MeLoId can recommend a few 
promising combinations of arguments for the
\verb|induct| method for a given goal.

Note that MeLoId's numerical evaluation on induction heuristics
is an instance of \textit{inductive reasoning}.
For example, even if all invocations of the \verb|induct| method in the AFPs are
compatible with a certain induction heuristic,
such information does not guarantee that the induction heuristic is always correct.
We can only state that the induction heuristic is \textit{probably} reliable.

\section{United Reasoning}

We envision \textit{united reasoning}, 
a fully automatic
inductive theorem prover embedded in Isabelle/HOL.
Upon success,
united reasoning combines the forces of
PSL's deductive reasoning,
PGT's abductive reasoning, and
MeLoId's inductive reasoning,
transforming PSL's depth-first search
into a best-first search,
so that we can automatically prove difficult inductive problems
that involve nested inductions and many conjecturing steps
while keeping the search space at a manageable size.

\begin{figure}
    \centering
    \includegraphics[width=1.0\linewidth]{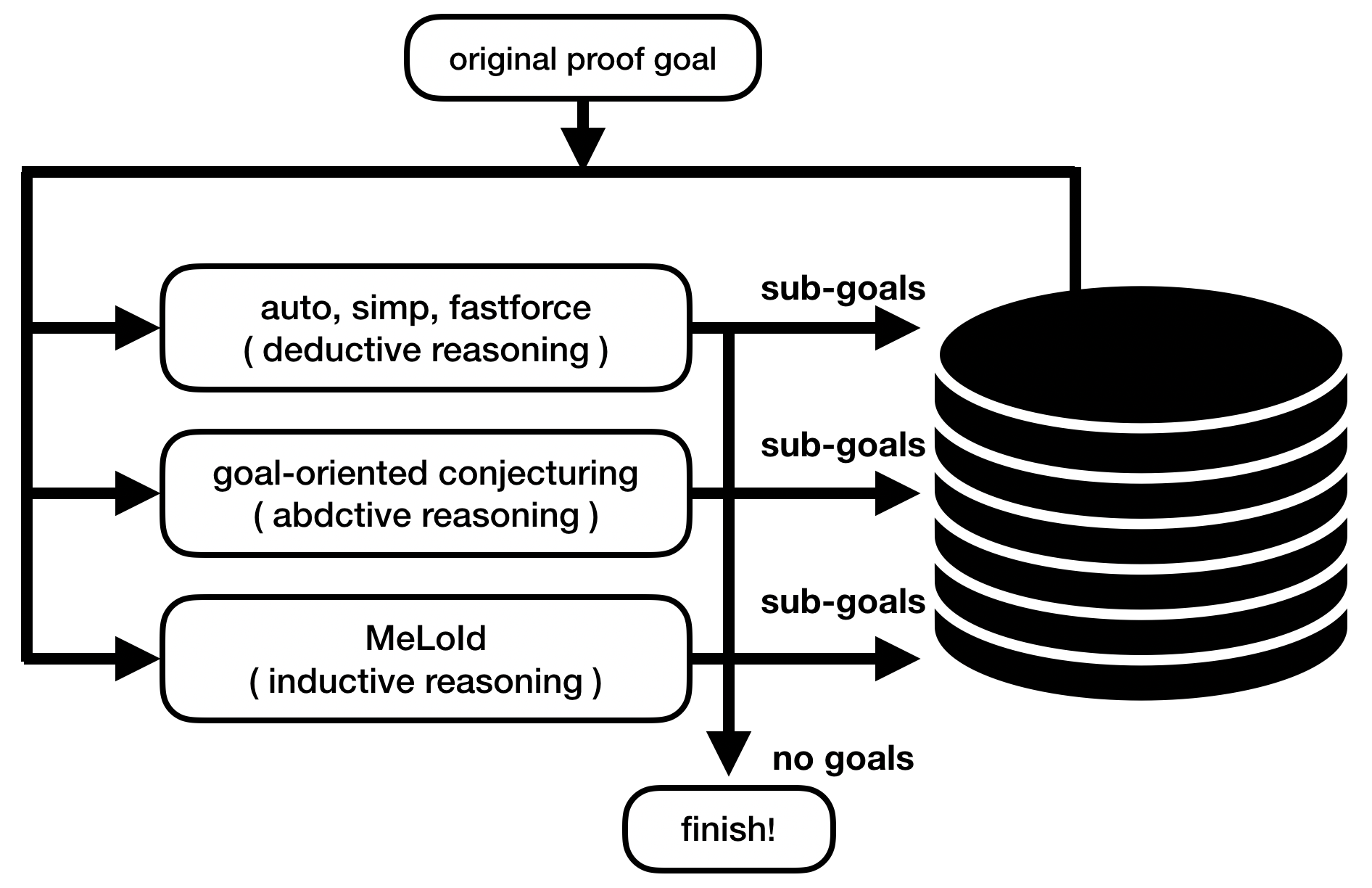}
    \caption{United Reasoning}
    \label{fig:ur}
\end{figure}

Figure \ref{fig:ur} illustrates the overall architecture
of united reasoning:
when the system receives a proof goal,
the system passes the goal to
three reasoning mechanisms:
deductive reasoning by Isabelle's
standard proof automation tools,
abductive reasoning by PGT's goal-oriented conjecturing,
and inductive reasoning with MeLoId.
If deductive reasoning discharges
the proof goal, the system 
stops working, 
printing the proof method, with which
it discharges the proof goal.
If there are remaining sub-goals,
united reasoning stores such sub-goals
in a priority queue.
Then, it keeps applying three reasoning mechanisms
to the most promising set of sub-goals,
which is stored at the top of the priority queue,
until the system discharges the original proof goal completely or the queue becomes empty.

Since the goal-oriented conjecturing
removes most irrelevant conjectures and
MeLoId is expected to give us a few
promising applications of the \verb|induct| method,
we hope that
the system can have a few number of the most promising sets of sub-goals near the top of the priority queue.

So far, it is unclear how we should prioritize the three reasoning mechanisms:
since existing proof corpora present only one proof for each theorem, 
a naive application of supervised learning does not seem to be a promising approach to
learning when to apply which school of reasoning.
We expect that the approach based on evolutionary computation to theorem proving \cite{evolution} may help us to attack this problem;
however, this still remains as our future work.

The most famous approach to automating proof by induction is called the waterfall model.
Compared to the original waterfall model, 
which ``\textit{uses no search}'' and
``\textit{is designed to make the right guess the first time,
and then pursue one goal with power and perseverance}'' \cite{waterfall},
we are designing unified reasoning with search in mind.
Unlike Moore, the creator of the waterfall model, 
we plan to make united reasoning
a search-based software
and place less importance on the speed of proof search
because we trace the proof search
using the writer monad transformer in Isabelle/ML \cite{constructor_class}
and produce efficient proof scripts upon successful
(potentially slow) proof search
as we did so with PSL.

%\begin{center}%\vspace{1cm}
%\includegraphics[width=1.0\linewidth]{acmart-master-4/samples/united_reasoning.png}
%\end{center} %\vspace{1cm}
\newpage
\begin{acks}
This work was supported by the European Regional Development Fund under the project 
AI \& Reasoning (reg. no.CZ.02.1.01/0.0/0.0/\\15\_003/0000466).
\end{acks}

\bibliographystyle{ACM-Reference-Format}
\bibliography{sample-base}

%%\appendix

\end{document}